\documentclass[10pt,twocolumn,letterpaper]{article}

\usepackage{cvpr}
\usepackage{times}
\usepackage{epsfig}
\usepackage{graphicx}
\usepackage{amsmath}
\usepackage{amssymb}
\usepackage{authblk}
\usepackage{pifont}
\usepackage{stfloats}
\usepackage{boldline,multirow} 
\usepackage{subcaption}
\usepackage{makecell}

\usepackage[breaklinks=true,bookmarks=false]{hyperref}

\cvprfinalcopy 

\def\cvprPaperID{****} 

\begin{document}

\title{Class-balanced Grouping and Sampling for Point Cloud 3D Object Detection} 

\author[1]{Benjin Zhu }
\author[1,2]{Zhengkai Jiang }
\author[3]{Xiangxin Zhou }
\author[1]{Zeming Li } 
\author[1]{Gang Yu } 
\affil[1]{Megvii Research} 
\affil[2]{Institute of Automation, Chinese Academy of Sciences}
\affil[3]{Tsinghua University}
\affil[ ]{\textit {\{zhubenjin, lizeming, yugang\}@megvii.com, zhengkai.jiang@nlpr.ia.ac.cn, xx-zhou16@mails.tsinghua.edu.cn}}


\maketitle

\begin{abstract}
    This report presents our method which wins the nuScenes 3D Detection Challenge \cite{nusc3ddet} held in Workshop on Autonomous Driving(WAD, CVPR 2019). Generally, we utilize sparse 3D convolution to extract rich semantic features, which are then fed into a class-balanced multi-head network to perform 3D object detection. To handle the severe class imbalance problem inherent in the autonomous driving scenarios, we design a class-balanced sampling and augmentation strategy to generate a more balanced data distribution. Furthermore, we propose a balanced grouping head to boost the performance for the categories with similar shapes.  Based on the Challenge results, our method outperforms the PointPillars  \cite{pointpillars} baseline by a large margin across all metrics, achieving state-of-the-art~(SOTA) detection performance on the nuScenes dataset. Code will be released at \href{https://github.com/poodarchu/Class-balanced-Grouping-and-Sampling-for-Point-Cloud-3D-Object-Detection}{CBGS}.
\end{abstract}

\section{Introduction}
Point cloud 3D object detection has recently received more and more attention and becomes an active research topic in 3D computer vision community since it has great potential for visual applications like autonomous driving and robots navigation. The KITTI dataset~\cite{Geiger2013IJRR} is the most widely used dataset in this task. Recently, NuTonomy releases the nuScenes dataset~\cite{nuscenes}, which greatly extends KITTI in dataset size, sensor modalities, categories, and annotation numbers. Compared to the KITTI 3D detection benchmark \cite{Geiger2012CVPR}, in which we need to locate and classify objects of 3 categories respectively, the nuScenes 3D Detection Challenge requires to detect 10 categories at the same time. Moreover, we need to estimate a set of attributes and object velocities for each object. Furthermore, the nuScenes dataset \cite{nuscenes} suffers from severe class imbalance issues. As shown in Figure \ref{fig:insnumcls}, instance distribution of categories in the nuScenes dataset is long-tailed, exhibiting an extreme imbalance in the number of examples between common and rare object classes. All the above challenges make the nuScenes 3D Detection Challenge more difficult, yet closer to real-world scenarios.

Existing 3D object detection methods have explored several ways to tackle 3D object detection task. Several works \cite{DBLP:journals/corr/ChenMWLX16,8594049, Liang_2018_ECCV,Yang_2018_CVPR,pointpillars} convert point cloud into bird-view format and apply 2D CNN to get 3D object detection results. Voxel-based methods \cite{7780463,Zhou_2018_CVPR,Yan_2018} convert point cloud into regular 3D voxels then apply 3D CNN or 3D sparse convolution \cite{DBLP:journals/corr/GrahamM17,DBLP:journals/corr/abs-1711-10275,DBLP:journals/corr/abs-1904-08755} to extract features for 3D object detection. Point-based Methods \cite{DBLP:journals/corr/abs-1711-08488,DBLP:journals/corr/abs-1711-10871} firstly utilize 2D detectors to obtain 2D boxes from the image, and then apply PointNet++ \cite{DBLP:journals/corr/QiSMG16,DBLP:journals/corr/QiYSG17} on the cropped point cloud to further estimate location, size and orientation of 3D objects. Methods taking advantage of both voxel-based and point-based methods like \cite{shi2019part,std2019yang,Shi_2019_CVPR} first use pointnet fashions to acquire high-quality proposals, then voxel-based methods is applied to obtain final predictions. However, most of above methods are performed on each single category respectively in order to achieve their highest performance. For example, the previous SOTA method PointPillars \cite{pointpillars} can only achieve very low performance on most of the rare categories(e.g., Bicycle).

 Multi-task Learning is another technique that we use in the challenge because the multi-category joint detection can be taken as a multi-task learning problem. Many works investigate how to adaptively set weights for the different task effectively. For example, MGDA \cite{DBLP:journals/corr/abs-1810-04650} takes multi-task learning as a multi-objective optimization problem. GradNorm \cite{DBLP:journals/corr/abs-1711-02257} uses gradient normalization strategies to balance loss of different tasks adaptively. Benefiting from multi-task learning, our method performs better when training all categories jointly than training each of them individually.

There are 3 tracks in the nuScenes 3D Detection Challenge: Lidar Track, Vision Track, and Open Track. Only lidar input is allowed in Lidar Track. Only camera input is allowed in Vision Track. External data or map data is not allowed in above two tracks. As for Open Track, any input is allowed. Besides, pre-training is allowed in all of the 3 tracks. We participate in the Lidar Track of the challenge. Final leaderboard can be found at \cite{nusc3ddet}. Finally, our contributions in this challenge can be concluded as follows:
\begin{itemize}
    \item We propose class-balanced sampling strategy to handle extreme imbalance issue in the nuScenes Dataset.
    \item We design a multi-group head network to make categories of similar shapes or sizes could benefit from each other, and categories of different shapes or sizes stop interfere with each other. 
    \item  Together with improvements on network architecture, loss function, and training procedure, our method achieves state-of-the-art performance on the challenging nuScenes Dataset \cite{nuscenes}.
\end{itemize}

We first introduce our methodology in Section \ref{sec:method}. Training details and network settings are presented in Section \ref{sec:training}. Results are shown in Section \ref{sec:results}. Finally we conduct conclusion in Section \ref{sec:conclusion}.

\bigskip
 \begin{table*}
 \begin{center}
     \begin{tabular}{c|c|c|c|c}     
     \Xhline{0.8pt}
     \textbf{Class} & \textbf{Instance Num} & \textbf{Sample Num} & \textbf{Instance Num After}& \textbf{Sample Num After} \\
    \Xhline{0.8pt}
     Car & 413318 & 27558 & 1962556 & 126811\\
     \hline
     Truck & 72815 & 20120 & 394195 & 104092\\
     \hline
     Bus & 13163 & 9156 & 70795 & 49745 \\
     \hline
     Trailer & 20701 & 7276 & 125003 & 45573 \\
     \hline
     Constr. Veh. & 11993 & 6770 & 82253 & 46710 \\
     \hline
     Pedestrian & 185847 & 22923 & 962123 & 110425 \\
     \hline
     Motocycle & 10109 & 6435 & 60925 & 38875\\
     \hline
     Bicycle & 9478 & 6263 & 58276 & 39301\\
     \hline
     Traffic Cone & 82362 & 12336 & 534692 & 73070\\
     \hline
     Barrier & 125095 & 9269 & 881469 & 60443 \\
    \hline
    Total & 944881 & 28130 & 5132287 & 128100 \\
    \Xhline{0.8pt}
 \end{tabular}
 \end{center}
 \caption{\textbf{Instance and sample distribution of training split before and after dataset sampling(DS Sampling).} Column \textbf{Instance Num} indicates instance number of each category. Column \textbf{Sample Num} indicates total sample numbers that a category appears in the training split. Column \textbf{Instance Num After} indicates instance number of each category after dataset sampling which expands the training set from 28130 to 128100 samples. Column \textbf{Sample Num After} is the same as column \textbf{Instance Num After}. Total number of samples indicates training dataset size, rather than the sum of all categories listed above, considering the fact that multiple categories can appear in the same point cloud sample.}
 \label{table:inssamnum}
 \end{table*}
 \bigskip
 


\section{Methodology} \label{sec:method}
Overall network architecture is presented in Figure \ref{figure:netarch}, which is mainly composed of 4 part: Input Module, 3D Feature Extractor, Region Proposal Network, and Multi-group Head network. Together with improvements on data augmentation, loss function, and training procedure, we not only make it perform 10 categories' 3D object detection, velocity and attribute prediction simultaneously, but also achieve better performance than perform each category's detection respectively.

In this section, we first introduce inputs and corresponding data augmentation strategies. Then the 3D Feature Extractor, Region Proposal Network, and Multi-group head network will be explained in detail. Finally, improvements on loss, training procedure as well as other tricks will be introduced.

\subsection{Input and Augmentation} \label{datasetaug}

The nuScenes dataset provides point cloud sweeps in $(x, y, z, intensity, ring index)$ format, each of them associated with a time-stamp. We follow the fashion of official nuScenes baseline \cite{nuscenes}  by accumulating 10 Lidia sweeps to form dense point cloud inputs. Specifically, our input is of $(x, y, z, intensity, \Delta t)$ format. $\Delta t$ is the time lag between each non-keyframe sweep regarding keyframe sweep, and $\Delta t$ ranges from 0s to 0.45s. We use grid size 0.1m, 0.1m, 0.2m in x, y, z axis respectively to convert the raw point cloud into voxel presentation. In each voxel, we take mean of all points in the same voxel to get final inputs to the network. No extra data normalization strategy is applied.

\begin{figure}
\begin{center}
\includegraphics[width=0.45 \textwidth]{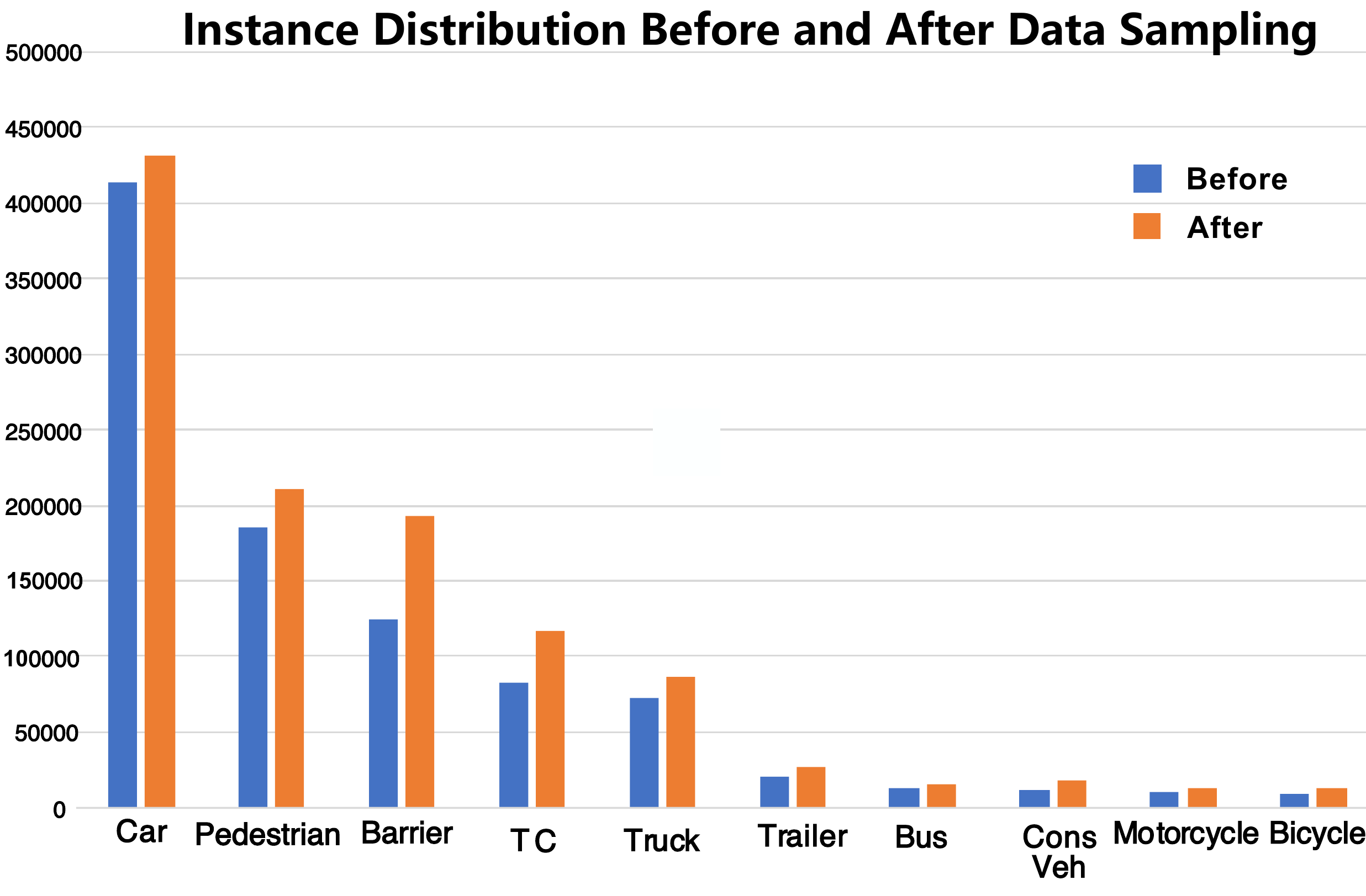}
\end{center}
\caption{\textbf{Class imbalance in the nuScenes Dataset.} 50\% categories account for only a small fraction of total annotations. Distribution of original Training Split is shown in blue. Distribution of sampled Training Split is shown is orange.}
\label{fig:insnumcls}
\end{figure}

As shown in Figure \ref{fig:insnumcls}, the nuScenes dataset \cite{nuscenes} has a severe class imbalance problem . Blue columns tell the original distribution of training split. To alleviate the severe class imbalance, we propose \textbf{DS Sampling}, which generates a smoother instance distribution as the orange columns indicate. To this end, like the sampling strategy used in the image classification task, we firstly duplicate samples of a category according to its fraction of all samples. The fewer a category's samples are, more samples of this category are duplicated to form the final training dataset.  More specifically, we first count total point cloud sample number that exists a specific category in the training split, then samples of all categories which are summed up to 128106 samples. Note that there exist duplicates because multiple objects of different categories can appear in one point cloud sample. Intuitively, to achieve a class-balanced dataset, all categories should have close proportions in the training split. So we randomly sample 10\% of 128106 (12810) point cloud samples for each category from the class-specific samples mentioned above. As a result, we expand the training set from 28130 samples to 128100 samples, which is about 4.5 times larger than the original dataset. To conclude, \textbf{DS Sampling} can be seen as improving the average density of rare classes in the training split. Apparently, \textbf{DS Sampling} could alleviate the imbalance problem effectively, as shown in orange columns in Figure \ref{fig:insnumcls}.

\begin{figure*}
\centering
\begin{subfigure}{.33\textwidth}
  \centering
  \includegraphics[width=.9\linewidth]{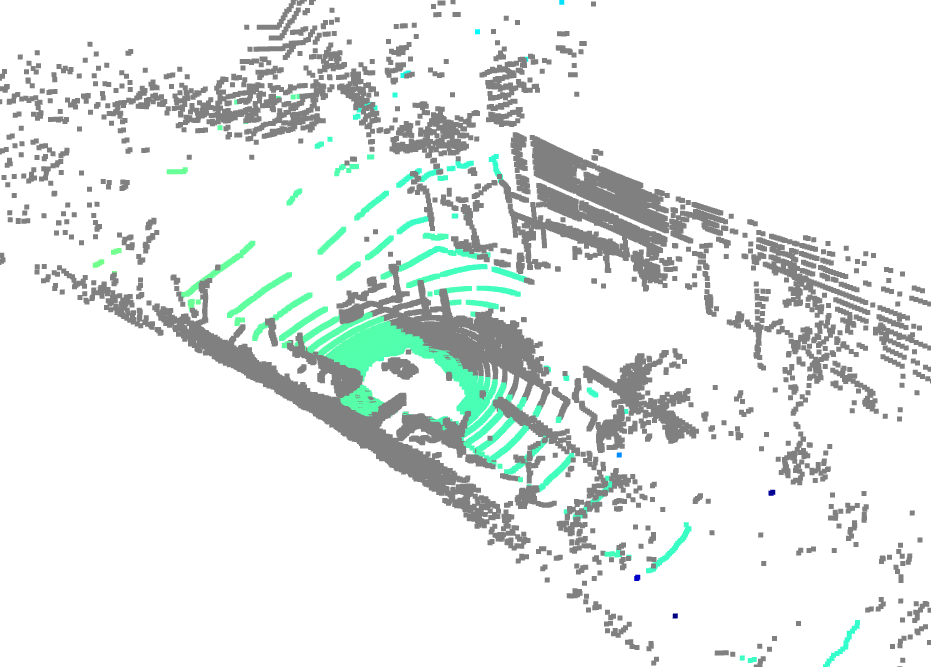}
  \label{fig:sub1}
\end{subfigure}%
\begin{subfigure}{.33\textwidth}
  \centering
  \includegraphics[width=.9\linewidth]{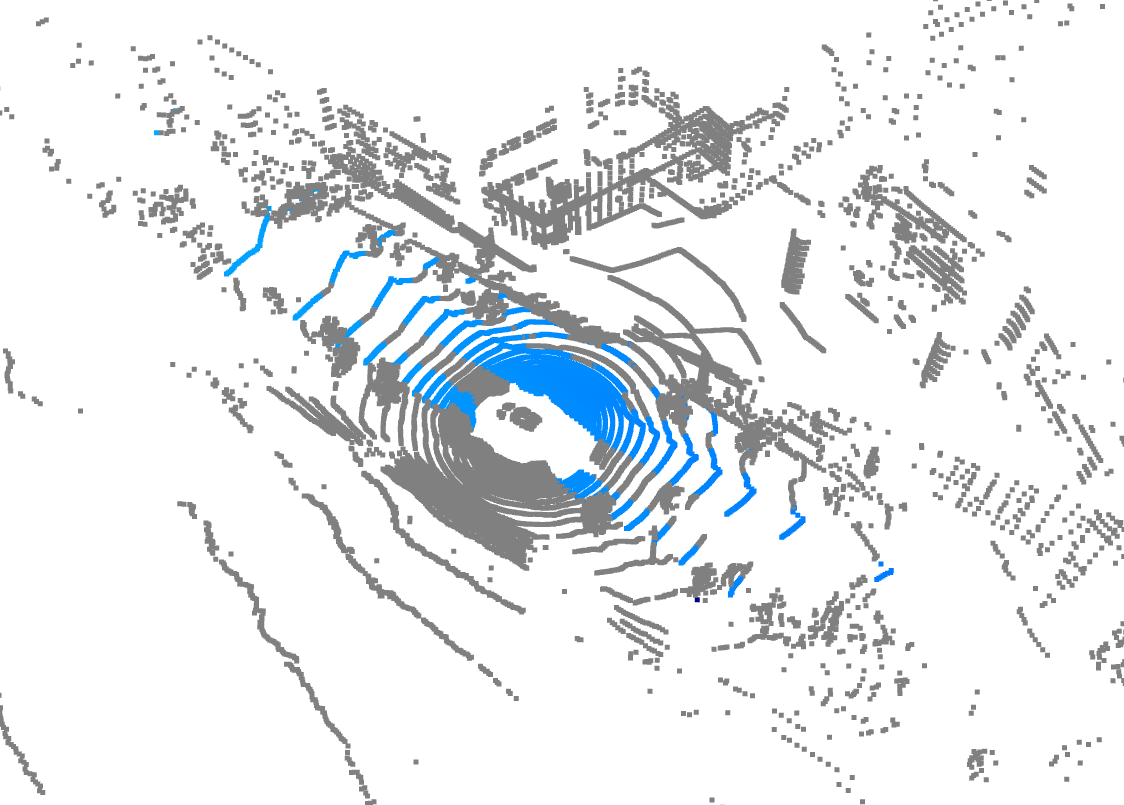}
  \label{fig:sub2}
\end{subfigure}
\begin{subfigure}{.33\textwidth}
  \centering
  \includegraphics[width=.9\linewidth]{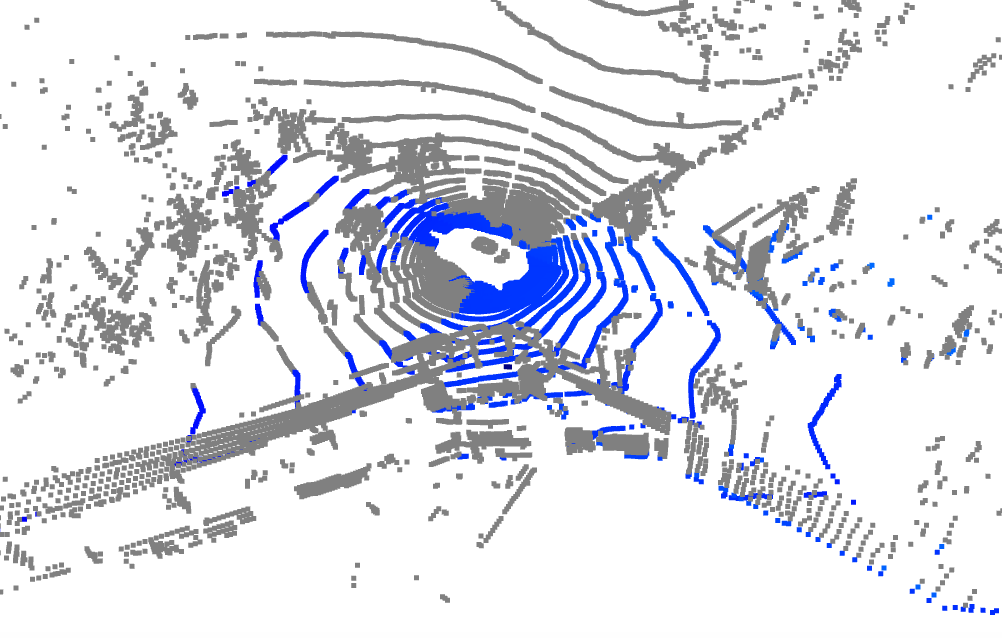}
  \label{fig:sub3}
\end{subfigure}
\caption{\textbf{Examples of ground plane detection result.} Points belonging to ground plane are shown in color, which can be formulated by $Ax + By + Cz + D = 0$. In average, the ground plane is about -1.82 meters along z axis. Open3D \cite{DBLP:journals/corr/abs-1801-09847} is used for visualization.}
\label{fig:gpdet}
\end{figure*}

Besides, we use \textbf{GT-AUG} strategy as proposed in SECOND \cite{Yan_2018} to sample ground truths from an annotation database, which is generated offline, and place those sampled boxes into another point cloud. Note that the ground plane location of point cloud sample needs to be computed before we could place object boxes properly. So we utilize the least square method and RANSAC \cite{Fischler1987Random} to estimate each sample's ground plane, which can be formulated as $Ax+By+Cz+D=0$. Examples of our ground plane detection module can be seen in Figure \ref{fig:gpdet}.

With the help of the above two strategies, we enable the model to perform better in all, especially tail classes, showing an obvious promoting effect on alleviating the problem of class imbalance.

\subsection{Network}

\begin{figure*}[t]
  \centering
  \includegraphics[width=0.9 \textwidth]{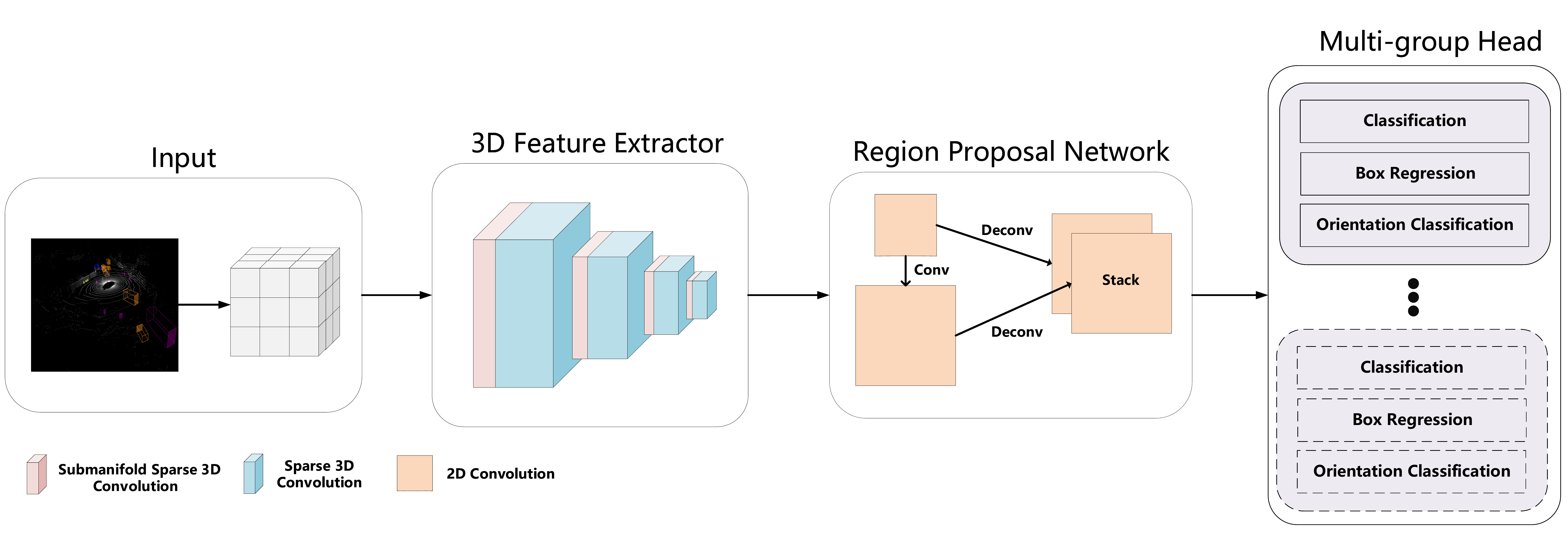} \\
  \caption{\textbf{Network Architecture.} 3D Feature Extractor is composed of submanifold and regular 3D sparse convolutions. Outputs of 3D Feature Extractor are of 16$\times$ downscale ratio, which are flatten along output axis and fed into following Region Proposal Network to generate 8$\times$ feature maps, followed by the multi-group head network to generate final predictions. Number of groups in head is set according to grouping specification.}
  \label{figure:netarch}
\end{figure*}

As Shown in Figure \ref{figure:netarch}, we use sparse 3D convolution with skip connections to build a resnet-like architecture for the 3D feature extractor network. For a $N \times C \times H \times W$ input tensor, the feature extractor outputs a $N \times l \times \frac{C}{m} \times \frac{H}{n} \times \frac{W}{n}$ feature map, $m, n$ is the downscale factor of z, x, y dimensions respectively, $l$ is output channel of 3D Feature Extractor's last layer. To make that 3D feature maps more suitable for the following Region Proposal Network and multi-group head which will be explained in detail in the next subsection, we reshape feature maps to $N \times \frac{C \times l}{m} \times \frac{H}{n} \times \frac{W}{n}$, then use a region proposal network like VoxelNet \cite{Zhou_2018_CVPR} to perform regular 2D convolution and deconvolution to further aggregate features and get higher resolution feature maps. Based on these feature maps the multi-group head network is thus able to detect objects of different categories efficiently and effectively.

\subsection{Class-balanced Grouping}
The intrinsic long-tail property poses a multitude of open challenges for object detection since the models will be largely dominated by those abundant head classes while degraded for many other tail classes. As shown in Figure \ref{fig:insnumcls}, for example, Car accounts for 43.7\% annotations of the whole dataset, which is 40 times the number of bicycle, making it difficult for a model to learn features of tail classes sufficiently. That is, if instance numbers of classes sharing a common head differ a lot, there is usually no data for the tail class at most time. As a result, the corresponding head, as the purple parts pictured in Figure \ref{figure:netarch}, will be dominated by the major classes, resulting in poor performance on rare classes. On the other hand, if we put classes of discrepant shapes or sizes together, regression target will have bigger inter-class variances, which will make classes of different shapes interfere with each other. That is why the performance trained with different shapes jointly is often lower than trained them individually. Our experiments prove that classes of similar shape or size are easier to learn from the same task.

Intuitively, classes of similar shapes or sizes can contribute to each other's performance when trained jointly because there are common features among those relative categories so that they can compensate for each other to achieve higher detection results together. To this end, we manually divide all categories into several groups following some principles. For a particular head in the Multi-group Head module, it only needs to recognize classes and locates objects belongs to classes of this group. There are mainly 2 principles which guide us split the 10 classes into several groups effectively:
\begin{itemize}
    \item \textbf{Classes of similar shapes or sizes should be grouped.} Classes of similar shapes often share many common attributes. For example, all vehicles look similar because they all have wheels, and look like a cube. Motorcycle and bicycle,  traffic cone and pedestrian also have a similar relation. By grouping classes of similar shape or size, we divide classification into two steps logically. Firstly the model recognizes 'superclasses', namely groups, then in each group, different classes share the same head. As a result, different groups learn to model different shape and size patterns, and in a specific group, the network is forced to learn the inter-class difference of similar shapes or sizes.
    \item \textbf{Instance numbers of different groups should be balanced properly.} We take into account that instance number of different groups should not vary greatly, which will make the learning process dominated by major classes. So we separate major classes from groups of similar shape or size. For example, Car, Truck and Construction Vehicle have similar shape and size, but Car will dominate the group if we put the 3 classes together, so we take Car as a single group, and put Truck and Construction Vehicle together as a group. In this way, we can control the weights of different groups to further alleviate the imbalance problem.
\end{itemize} 

Guided by the above two principles, in the final settings we split 10 classes into 6 groups: (Car), (Truck, Construction Vehicle), (Bus, Trailer), (Barrier), (Motorcycle, Bicycle), (Pedestrian, Traffic Cone). According to our ablation study as shown in Table \ref{table:ablation}, the class-balanced grouping contributes the most to the final result.

\begin{table*}[t]
    \begin{center}
    \begin{tabular} {c|c|c|c|c|c|c|c|c|c|c}
        \Xhline{0.8pt}
        & \textbf{Modality} & \textbf{Map} & \textbf{External} & \textbf{mAP} & \textbf{mATE} & \textbf{mASE} & \textbf{mAOE} & \textbf{mAVE} & \textbf{mAAE} & \textbf{NDS} \\
        \Xhline{0.8pt}
        Point Pillars \cite{pointpillars} & Lidar & $\mathcal{\times}$ & $\mathcal{\times}$ & 30.5 & 0.517 & 0.290 & 0.500 & 0.316 & 0.368 & 45.3 \\
        \hline
        BRAVE \cite{nusc3ddet} & Lidar & $\mathcal{\times}$ & $\mathcal{\times}$ &  32.4 & 0.400 & 0.249 & 0.763 & 0.272 & \textbf{0.090} & 48.4 \\
        \hline
        Tolist \cite{nusc3ddet} & Lidar & $\mathcal{\times}$ & $\mathcal{\times}$ & 42.0 & 0.364 & 0.255 & 0.438 & 0.270 & 0.319 & 54.5 \\
        \hline
        MEGVII(Ours) & Lidar & $\mathcal{\times}$ & $\mathcal{\times}$ & \textbf{52.8} & \textbf{0.300} & \textbf{0.247} & \textbf{0.380} & \textbf{0.245} & 0.140 & \textbf{63.3} \\
        \Xhline{0.8pt}
    \end{tabular}
    \end{center}
    \caption{\textbf{Overall performance.} BRAVE and Tolist are the other top three teams. Our method achieves the best performance on all but mAAE metrics.}
    \label{table:overall}
\end{table*}

\begin{table*}[t]
    \begin{center}
    \begin{tabular} {c|c|c|c|c|c|c|c|c|c|c|c}
        \Xhline{0.8pt}
        & Car & Ped & Bus & Barrier & TC & Truck & Trailer & Moto & Cons. Veh. & Bicycle & Mean \\
        \hline
        Point Pillars \cite{pointpillars} & 70.5 & 59.9 & 34.4 & 33.2 & 29.6 & 25.0 & 16.7 & 20.0 & 4.50 & 1.60 & 29.5\\
        \hline
        MEGVII(Ours) & \textbf{81.1} & \textbf{80.1} & 5\textbf{4.9} & \textbf{65.7} & \textbf{70.9} & \textbf{48.5} & \textbf{42.9} & \textbf{51.5} & \textbf{10.5} & \textbf{22.3} & \textbf{52.8} \\
        \Xhline{0.8pt}
    \end{tabular}
    \end{center}
    \label{table:percls}
    \caption{\textbf{mAP by Categories compared to PointPillars.} Our method shows more competitive and balanced performance on tail classes. For example, Bicycle is improved by 14 times. Motorcycle, Construction Vehicle(Cons. Veh.), Trailer, and Traffic Cone(TC) are improved by more than 2 times.}
\end{table*}

\begin{table*}[t]
    \begin{center}
        \begin{tabular} {c|c|c|c|c|c|c|c|c|c}
            \Xhline{0.8pt}
            GT-AUG & DB Sampling & Multi-head & Res-Encoder & SE & Heavier Head & WS & Hi-res & mAP & NDS \\
            \Xhline{0.8pt}
            $\times$ & $\times$ & $\times$ & $\times$ & $\times$ & $\times$ & $\times$ & $\times$ & 35.68 & 45.17 \\
            \checkmark & $\times$ & $\times$ & $\times$ & $\times$ & $\times$ & $\times$ & $\times$ & 37.69 & 53.66 \\
            \checkmark & \checkmark & $\times$ & $\times$ & $\times$ & $\times$& $\times$ & $\times$ & 42.64 & 56.66 \\
            \checkmark & \checkmark & \checkmark &$\times$  & $\times$ & $\times$ & $\times$ & $\times$ & 44.86 & 58.13 \\
            \checkmark & \checkmark & \checkmark & \checkmark & $\times$ & $\times$ & $\times$ & $\times$ & 48.64 & 60.08 \\
            \checkmark & \checkmark & \checkmark & \checkmark & \checkmark &$\times$ &$\times$  & $\times$  &48.14 & 59.66 \\
            \checkmark & \checkmark & \checkmark & \checkmark & \checkmark & \checkmark &  $\times$&$\times$ & 49.55 & 60.20 \\
            \checkmark & \checkmark & \checkmark & \checkmark & \checkmark & \checkmark & \checkmark &$\times$ & 49.43 & 60.56 \\
            \checkmark & \checkmark & \checkmark & \checkmark & \checkmark & \checkmark & \checkmark & \checkmark & 51.44 & 62.56 \\
            \Xhline{0.8pt}
        \end{tabular}
    \end{center}
    \caption{\textbf{Ablation studies for different components used in our method on Validation Split.} Database Sampling and Res-Encoder contribute the most to mAP.}
        \label{table:ablation}
\end{table*}

\subsection{Loss Function}
Apart from regular classification and bounding box regression branch required by 3D object detection, we add an orientation classification branch as proposed in SECOND \cite{Yan_2018}. It's important to point out that most of the object boxes are parallel or perpendicular to LiDAR coordinates axis according to our statistics. So if orientation classification is applied as it is in SECOND, it turns out the mAOE is very high for the fact that many predicted bounding boxes' orientation are just opposite to ground truth. So we add an offset to orientation classification targets to dismiss orientation ambiguity. As for velocity estimation, regression without normalization can achieve the best performance compared to adding extra normalization operations. 

We use anchors to reduce learning difficulty through import prior knowledge. Anchors are configured as VoxelNet \cite{Zhou_2018_CVPR}. That is, anchors of different classes have different height and width configuration which are determined by class means values. There is 1 size configuration with 2 different directions for a category. For velocities, the anchor is set to 0 in both x and y axis. Objects are moving along the ground so we do not need to estimate velocity in the z axis.

In each group, we use weighted Focal Loss for classification, the smooth-l1 loss for $x, y, z, l, w, h, yaw, v_x, v_y$ regression, and softmax cross-entropy loss for orientation classification. We do not add attribute estimation because its results are not comparable to just applying each category's most common attribute. We further improve attribute estimation by taking velocity into account. For example, most bicycles are $without\_rider$, but if the model predicts a bicycle's velocity is above a threshold, there should be riders so we change corresponding bicycle's attribute to $with\_rider$.

The Multi-group head is taken as a multi-task learning procedure in our experiments. We use \textbf{Uniform Scaling} to configure weights of different branches.

\subsection{Other Improvements}

Apart from the above improvements, we find that SENet \cite{Hu_2018_CVPR}, Weight Standardization \cite{DBLP:journals/corr/abs-1903-10520} can also help in the detection task when used properly. Besides, if we use a heavier head network, performance can still be improved. In our final submission, we ensemble several models of multiple scales to achieve our best performance: mAP 53.2\%, NDS 63.78\% on validation split.

\section{Training Details} \label{sec:training}

In this section, we explain the implementation details of the data augmentation, training procedure and method itself. Our method is implemented in PyTorch \cite{paszke2017automatic}. All experiments are trained using NVIDIA 2080Ti distributedly with synchronized batch normalization support.

For this task, we consider point cloud within the range of [-50.4, 50.4] $\times$ [-51.2, 51.2] $\times$ [-5, 3] meters in X, Y, Z axis respectively. We choose a voxel size of $s_x$ = 0.1, $s_y$ = 0.1, $s_z$ = 0.2 meters, which leads to a 1008 $\times$ 1024 $\times$ 40 voxels. Max points number allowed in a voxel is set to 10. For using 10 sweeps(1 keyframe + 9 preceeding non-keyframes), max number of non-empty voxels is 60000. 

During training, we conduct data augmentation of random flip in the x-axis, scaling with a scale factor sampled from [0.95, 1.05], rotation around Z axis between [-0.3925, 0.3925] rads and translation in range [0.2, 0.2, 0.2] m in all axis. For GT-AUG, we first filter out ground truth boxes with less than 5 points inside, then randomly select and paste ground truth boxes of different classes using different magnitude on the ground plane as shown in Table \ref{table:gtaug}.

\subsection{Training Procedure}
We use adamW \cite{DBLP:journals/corr/abs-1711-05101} optimizer together with one-cycle policy \cite{OneCyclePolicySylvainGugger} with LR max 0.04, division factor 10, momentum ranges from 0.95 to 0.85, fixed weight decay 0.01 to achieved super convergence. With batch size 5, the model is trained for 20 epochs. During inference, top 1000 proposals are kept in each group, then NMS with score threshold 0.1 and IoU threshold 0.2 is applied. Max number of boxes allowed in each group after NMS is 80.

\begin{table*}[t] 
\begin{center}    
\begin{tabular}{c|c|c|c|c|c|c|c|c|c|c}
    \Xhline{0.8pt}
    \textbf{Category} & Car & Truck & Bus & Trailer & Cons. Veh. & Traffic Cone & Barrier & Bicycle & Motorcycle & Pedestrian \\
    \Xhline{0.8pt}

    \textbf{Magnitude} & 2 & 3 & 7 & 4 & 6 & 2 & 6 & 6 & 2 & 2 \\
    \Xhline{0.8pt}
\end{tabular}
\end{center}
\caption{\textbf{GT-AUG magnitudes of different categories.} For each category, the magnitude means number of instances placed into a point cloud sample.}
\label{table:gtaug}
\end{table*}

\begin{figure*}[t]
\centering
\begin{subfigure}[b]{.33\linewidth}
\includegraphics[width=\linewidth]{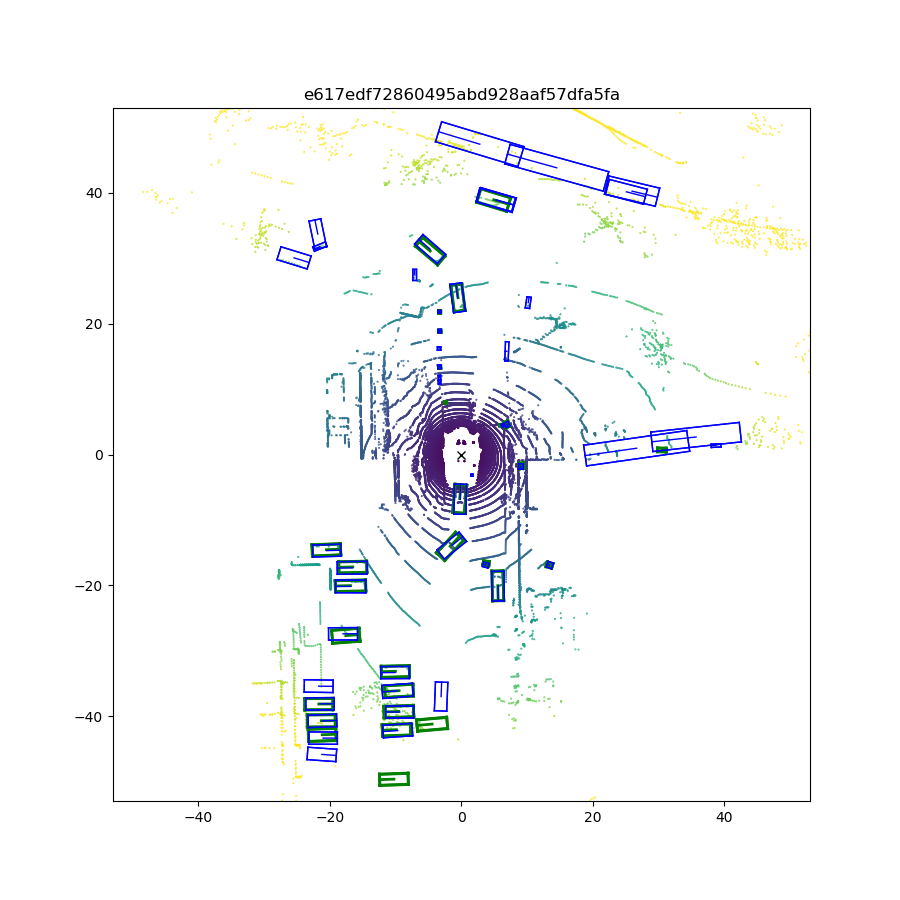}
\end{subfigure}
\begin{subfigure}[b]{.33\linewidth}
\includegraphics[width=\linewidth]{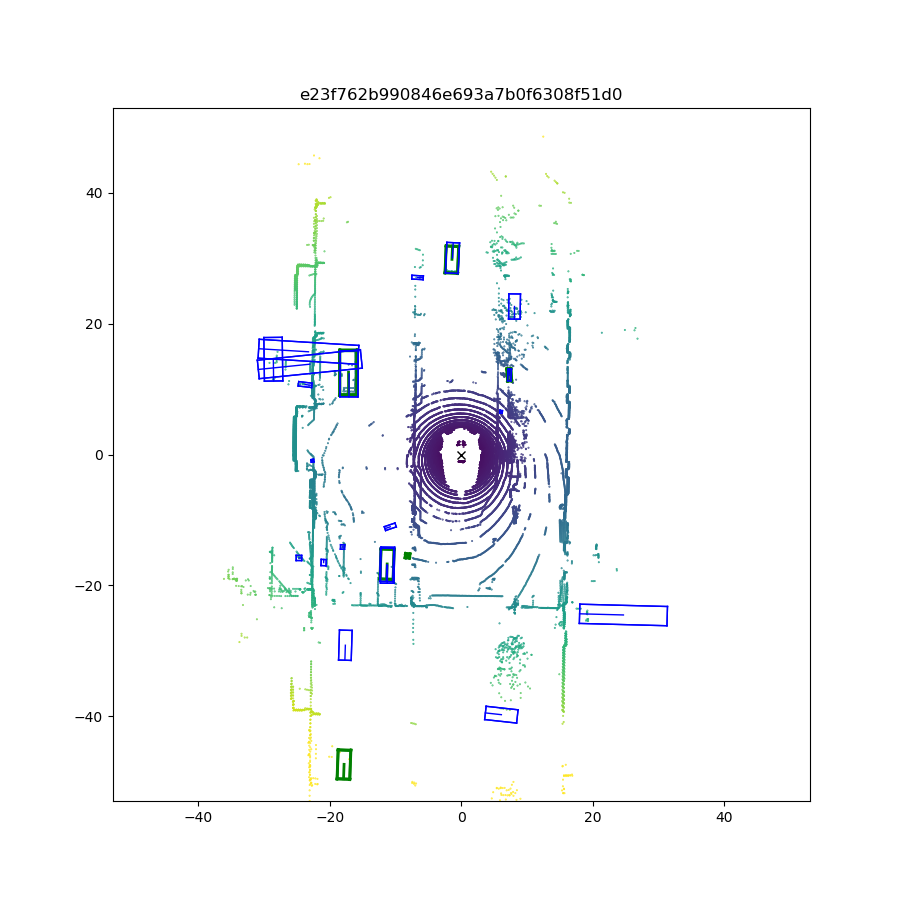}
\end{subfigure}
\begin{subfigure}[b]{.33\linewidth}
\includegraphics[width=\linewidth]{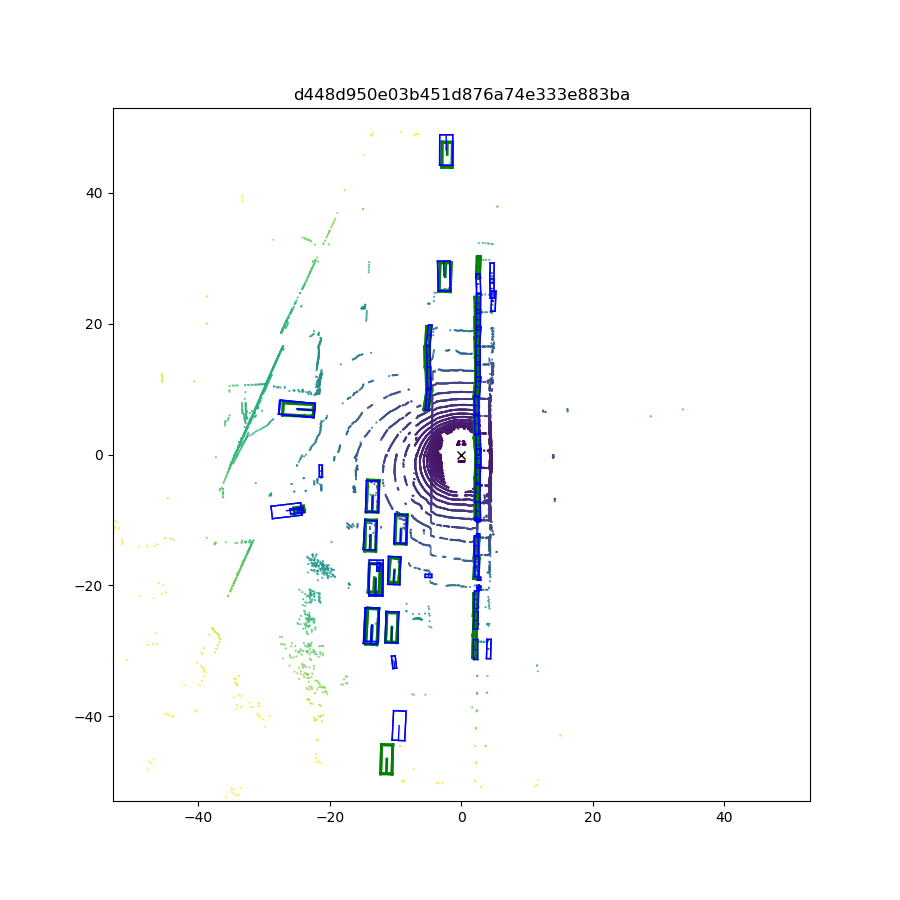}
\end{subfigure}
\begin{subfigure}[b]{.33\linewidth}
\includegraphics[width=\linewidth]{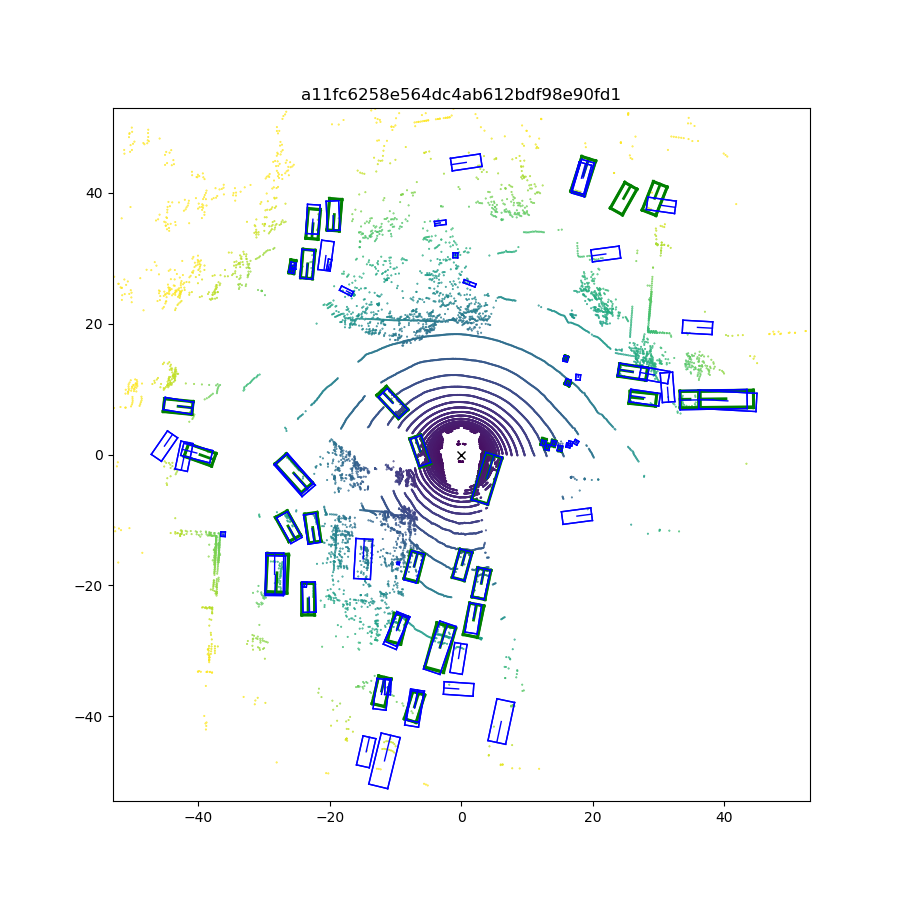}
\end{subfigure}
\begin{subfigure}[b]{.33\linewidth}
\includegraphics[width=\linewidth]{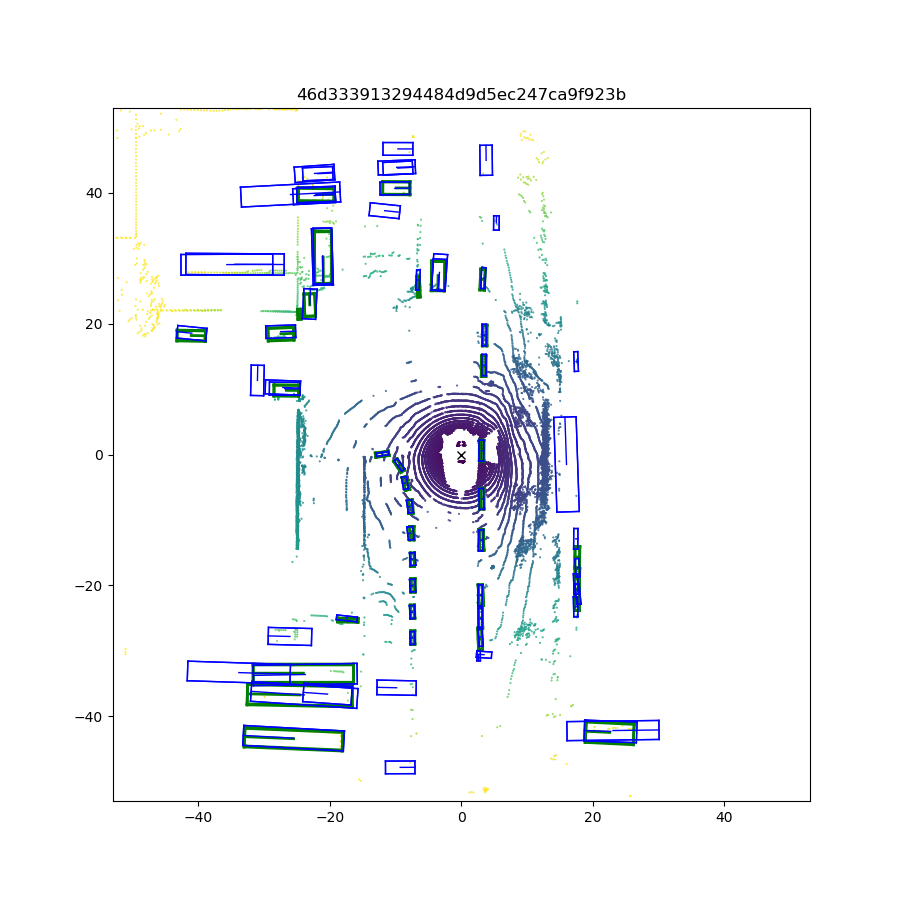}
\end{subfigure}
\begin{subfigure}[b]{.33\linewidth}
\includegraphics[width=\linewidth]{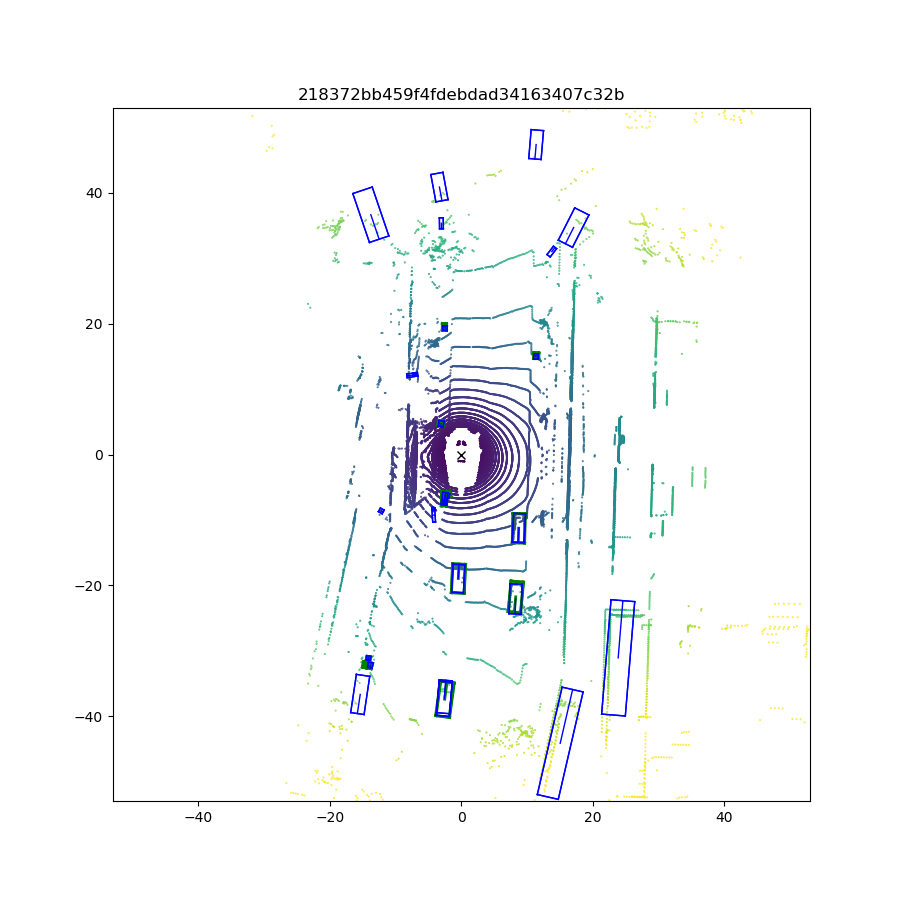}
\end{subfigure}
\begin{subfigure}[b]{.33\linewidth}
\includegraphics[width=\linewidth]{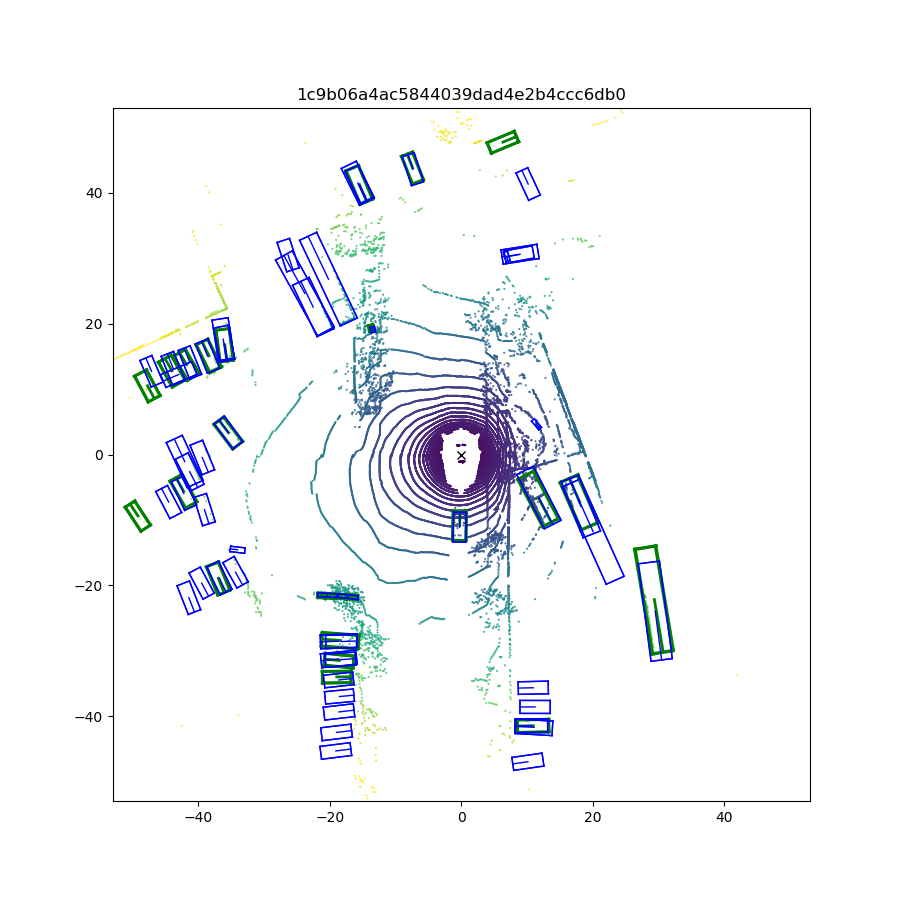}
\end{subfigure}
\begin{subfigure}[b]{.33\linewidth}
\includegraphics[width=\linewidth]{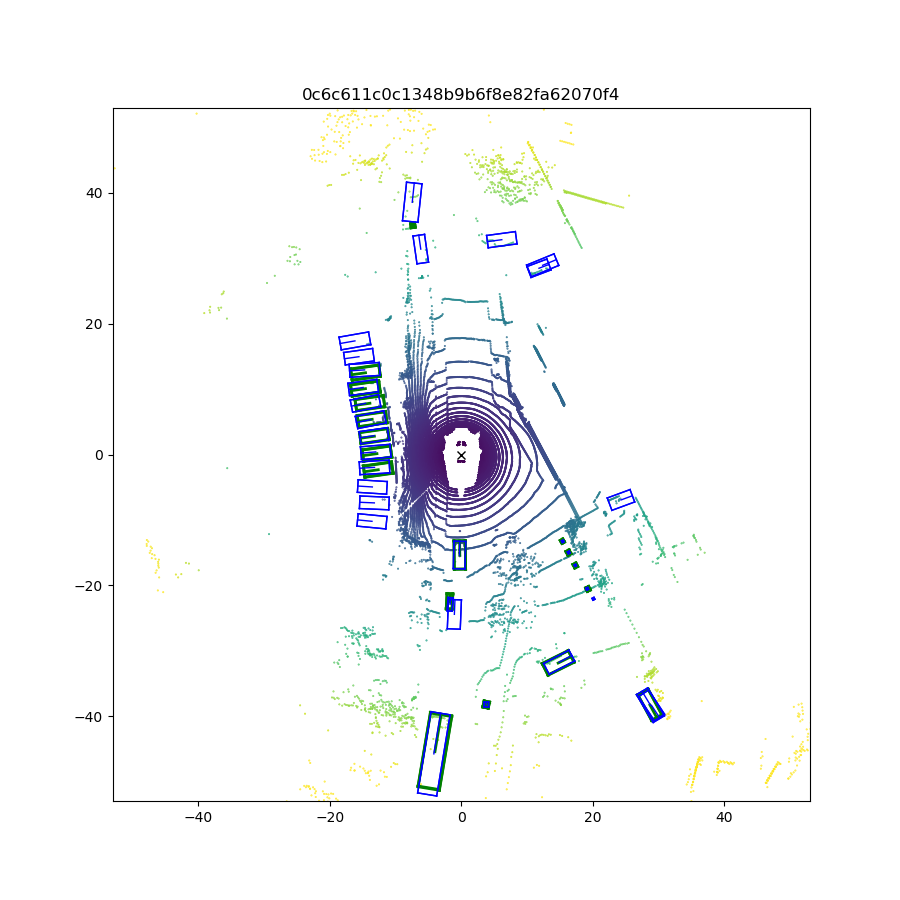}
\end{subfigure}
\begin{subfigure}[b]{.33\linewidth}
\includegraphics[width=\linewidth]{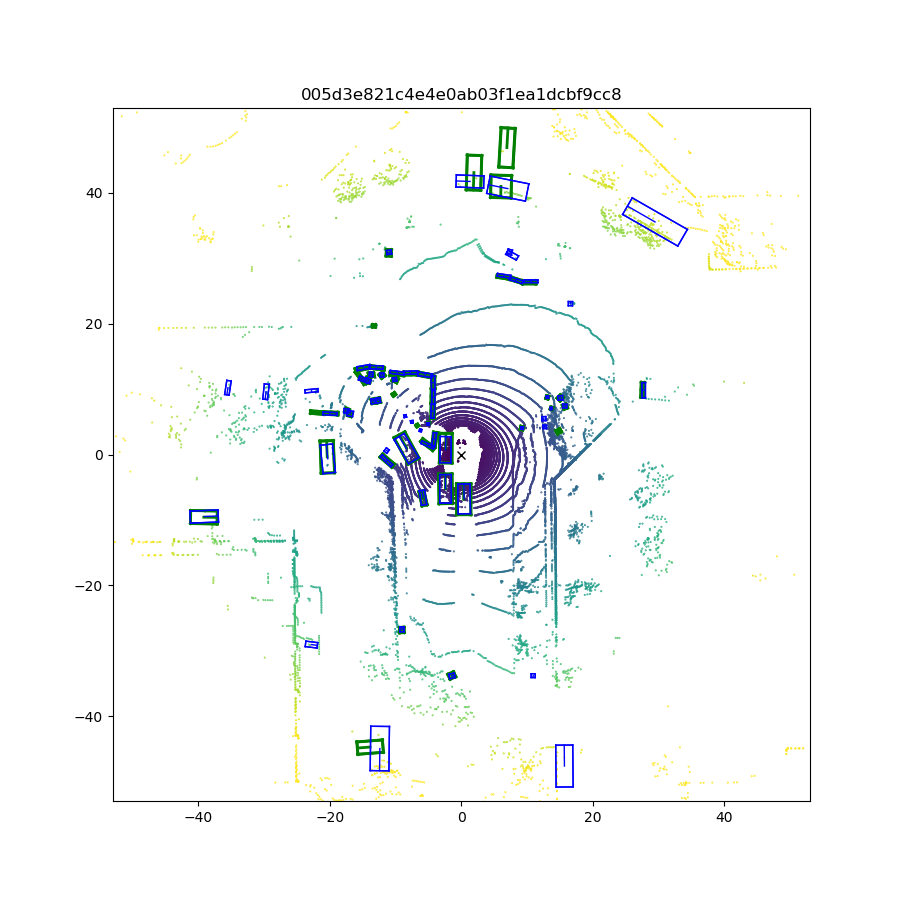}
\end{subfigure}
\caption{\textbf{Examples of detection results in validation split.} Ground truth annotations are in green and detection results are in blue. Detection results come from a model with 51.9\% mAP and 62.5\% NDS. The token on top of each point cloud bird view image is its corresponding sample data token.}
\label{fig:examplesbev}
\end{figure*}

\subsection{Network Details}

For the 3D feature extractor, we use 16, 32, 64, 128 layers of sparse 3D convolution respectively for each block. As used in \cite{DBLP:journals/corr/GrahamM17}, submanifold sparse convolution is used when we downsample the feature map. In other conditions, regular sparse convolution is applied. For the region proposal module, we use 128 and 256 layers respectively for downscale ratio 16$\times$ and 8$\times$ layers. In each head, we apply 1 $\times$ 1 Conv to get final predictions. To achieve a heavier head, we first use one layer 3 $\times$ 3 Conv to reduce channels by $\frac{1}{8}$, then use a 1 $\times$ 1 Conv layer to get final predictions. Batch Normalization \cite{DBLP:journals/corr/IoffeS15} is used for all but the last layer.

Anchors of different categories are set according to their mean height and width, with different threshold when assigning class labels. For categories of sufficient annotations, we set the positive area threshold to 0.6, for those categories with fewer annotations we set the threshold to 0.4.

We use the default setting of focal loss in the original paper. For $x, y, z, l, w, h, yaw, v_x, v_y$ regression, we use 0.2 for velocity prediction and the others are set to 1.0 to achieve a balanced and stable training process.

\section{Results} \label{sec:results}
In this section we report our results in detail. We also investigate contributions of each module to the final result in Table \ref{table:ablation}.

As shown in Table \ref{table:overall}, our method surpasses official PointPillars \cite{pointpillars} baseline by 73.1\%. More specifically, our method shows better performance in all categories, especially in long-tail classes like Bicycle, Motorcycle, Bus, and Trailer. Moreover, our method achieves less error in translation(mATE), scale(mASE), orientation(mAOE), velocity(mAVE) and attribute(mAAE). Examples of detection results can be seen in Figure \ref{fig:examplesbev}, our method generates reliable detection results on all categories. The edge with a line attached in the bounding box indicates the vehicle's front.

\section{Conclusion} \label{sec:conclusion}

In this report, we present our method and results on the newly-released large scale nuScenes Dataset, which poses more challenges, such as class imbalance, than KITTI on the 3D Object Detection task. With carefully-designed strategies in solving class imbalance, multi-class joint detection through data, network and learning objective, we achieve the best result in the WAD challenge. However, there are still a few methods that report their results on the nuScenes Dataset, so we will release our code, hopefully, it can facilitate people's research on this topic.

{\small

}

\end{document}